\pdfoutput=1

\documentclass[11pt]{article}

\usepackage[]{EMNLP2023}

\usepackage{times}
\usepackage{latexsym}

\usepackage{paralist}
\usepackage[T1]{fontenc}

\usepackage[utf8]{inputenc}

\usepackage{microtype}

\usepackage{inconsolata}

\usepackage{amsmath,amssymb,amsfonts}
\usepackage{url,enumitem}
\usepackage{booktabs}
\usepackage{xspace}
\usepackage{graphicx}
\usepackage{subcaption}
\usepackage{comment}
\usepackage[normalem]{ulem}
\usepackage{mathtools}
\usepackage{svg}
\usepackage{algorithm,algorithmicx,algpseudocode}
\usepackage{dsfont}

\usepackage{listings}
\newcommand{\Sref}[1]{\S\ref{#1}}

\DeclareMathOperator{\argmax}{argmax}

\newcommand{\methodname}[1]{\textsc{Ssd-LM}}
\newcommand{\newmethodname}[1]{\textsc{Ssd-2}}
\newcommand{\dolly}[1]{\textsc{dolly}}

\newcommand{\ignore}[1]{}

\let\orgautoref\autoref

\renewcommand{\autoref}[1]{\def\equationautorefname{Eq.}\orgautoref{#1}}

\title{
David helps Goliath: Inference-Time Collaboration Between\\ Small Specialized and Large General Diffusion LMs
}

\author{
  Xiaochuang Han$^{\spadesuit \diamondsuit}$ \quad \quad Sachin Kumar$^\clubsuit$ 
  \\
  \textbf{Yulia Tsvetkov}$^\spadesuit$ \quad \quad \quad \quad \textbf{Marjan Ghazvininejad}$^{\diamondsuit}$\\
  $^\spadesuit$University of Washington \qquad $^\diamondsuit$Meta AI \qquad $^\clubsuit$Carnegie Mellon University \\
        {\tt \{xhan77, yuliats\}@cs.washington.edu$^\spadesuit$ \quad sachink@cs.cmu.edu$^\clubsuit$ \quad ghazvini@meta.com$^{\diamondsuit}$} 
}

\begin{document}
\maketitle

\begin{abstract} 

Diffusion-based language models are emerging as a promising alternative to autoregressive LMs: they approach the competence of autoregressive LMs while offering nuanced controllability at inference time. 
While autoregressive LMs have benefited immensely from scaling and instruction-based learning, existing studies of diffusion LMs have been conducted on a smaller scale.
Starting with a recently proposed diffusion model \methodname{}, in this work we first explore methods to scale it from 0.4B to 13B parameters, proposing techniques to improve its training and inference efficiency, and to finetune the model to follow instructions. 
Armed with a more powerful, general purpose diffusion LM, we introduce the primary contribution of this work  -- \newmethodname{} -- an approach to easily ensemble at inference time a large general-purpose diffusion LM with smaller, but specialized and contextualized diffusion LMs.  
We show that \newmethodname{} facilitates novel ensembles with 100x smaller models that can be customized and deployed by individual users. We find that compared to autoregressive models, the collaboration between diffusion LMs is more effective, leading to higher-quality model responses due to their ability to dynamically incorporate bi-directional contexts.

\end{abstract}

\begin{figure}[ht]
    \centering
    \includegraphics[width=0.42\textwidth]{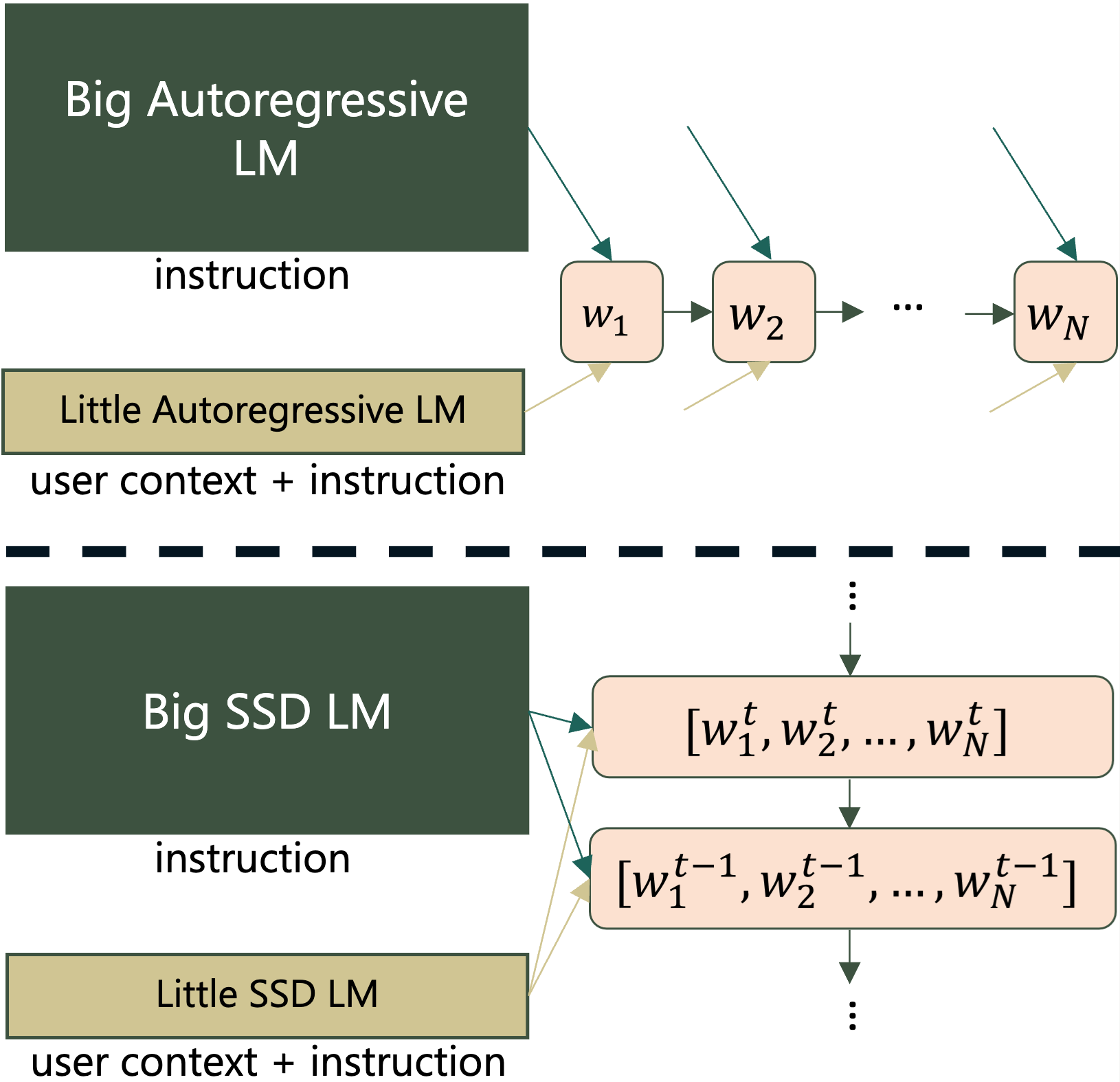}
    \caption{
    \emph{Inference-time collaboration} between a large general model and a small user model that incorporates user-specified knowledge. 
    The collaboration between autoregressive models performs decoding token-by-token, while the collaboration between diffusion models refines a block of generated tokens iteratively with bi-directional contexts (\Sref{sec:fusion}). 
    }
    \label{fig:fusion_fig}
\vspace{-1.0em}
\end{figure}

\section{Introduction}

Following the footsteps of diffusion-based generative models for continuously valued data such as images, audio, and video  \citep{ho2020denoising,Kong2021DiffWaveAV,Ho2022VideoDM}, recent works have attempted to replicate these successes on discrete text data \citep{Austin2021StructuredDD,Li2022DiffusionLMIC,Han2022SSDLMSS,Strudel2022SelfconditionedED,Dieleman2022ContinuousDF}. Several studies have shown that diffusion-based language models (LMs) perform competitively to their autoregressive counterparts, 
and even surpass them at post-hoc
controllable text generation \citep{Li2022DiffusionLMIC,Han2022SSDLMSS}. 

Meanwhile, autoregressive language models \citep{Brown2020LanguageMA,Touvron2023LLaMAOA} have emerged as general-purpose solutions capable of holding conversations with humans and solving tasks by following  instructions~\citep{Ouyang2022TrainingLM,Wang2022SuperNaturalInstructionsGV,Longpre2023TheFC,alpaca,vicuna2023,Xu2023BaizeAO}. 
Their abilities are primarily due to two factors: scaling the model parameters as well as pretraining datasets, and instruction finetuning with carefully curated datasets~\citep{Ouyang2022TrainingLM}. 

However, as the models become increasingly immense and proprietary, it is difficult for individual users to customize the system with their own data (e.g., specialized knowledge) due to cost or privacy reasons (\Sref{sec:fusion}). 
A primary contribution of this work is to illustrate a novel setup of \emph{inference-time collaboration} between LMs and show a unique advantage of diffusion LMs compared to autoregressive models in this scenario. 
With diffusion models' iterative generation design over a span of bi-directional contexts, multiple diffusion LMs with different capabilities can be easily ensembled at the sequence level at test time, leveraging advantages of each LM in the ensemble. 

As a preliminary to our experiments, we first present an exploratory study to scale and incorporate instruction-following and conversational capabilities in diffusion-based LMs. 
We introduce \newmethodname{}, an improved version of recently introduced simplex-based diffusion LM \methodname{}~\citep{Han2022SSDLMSS} proposing several modifications to its training and inference procedures. 
We incorporate these improvements in 
scaling \newmethodname{} to 13B parameters, up from 0.4B in \methodname{}. 
We show that similarly to autoregressive LMs, by finetuning with curated instruction datasets, \newmethodname{} is well-suited to follow chat-style instructions. 


We then present our main case study highlighting the setup of inference-time collaboration: we augment a general-purpose large \newmethodname{} model with 13B parameters with a 100x smaller, user-accessible model. This setup allows incorporating user-provided knowledge into the generation process without directly inputting it into the large model (which can be undesirable due to cost or privacy reasons).  
We show that \newmethodname{}'s instruction finetuned model is substantially more effective at this collaboration than the autoregressive baselines, leveraging bi-directional contexts in the ensemble.

\section{Background} 
Semi-autoregressive simplex-based diffusion LM (\methodname{}) is trained to generate text in blocks of tokens by performing diffusion in the simplex space of the model vocabulary \citep{Han2022SSDLMSS}. 
For text continuation tasks, it has shown competitive performance against autoregressive models~\citep[e.g., GPT-2;][]{Radford2019LanguageMA} when trained with a similar number of model parameters and pretraining data. 
Furthermore, it naturally enables post-hoc control in generated text using off-the-shelf classifiers, outperforming prior approaches to controlling autoregressive models.
Below we briefly overview the training and decoding algorithm of \methodname{}. 

\paragraph{Training} 
The core idea behind the training of diffusion models \citep{ho2020denoising} is to add a series of progressive noise to the input data representations and learn a model to reverse this process, reconstructing the original data at different noise levels. 
Assume we have a sequence of tokens $\{ w^0, \ldots, w^{c-1}, w^{c}, \ldots, w^{c+B-1} \}$, where we condition on a context of length c, $\{ w^0, \ldots, w^{c-1} \}$ (or $\boldsymbol{w}^{<c}$), and learn to generate the subsequent block of text $\{ w^{c}, \ldots, w^{c+B-1} \}$ (or $\boldsymbol{w}^{c:c+B}$) containing $B$ tokens. 
In \methodname{}, a progressive Gaussian noise is added to the block of text $\boldsymbol{w}^{c:c+B}$.  
\begin{align*}
    \Tilde{\boldsymbol{w}}_0^{c:c+B} &= \operatorname{logits-initialization}(\boldsymbol{w}^{c:c+B}) \\
    \Tilde{\boldsymbol{w}}_t^{c:c+B} &= \sqrt{\Bar{\alpha}_t} \Tilde{\boldsymbol{w}}_0^{c:c+B} + \sqrt{1-\Bar{\alpha}_t} \boldsymbol{\epsilon} 
\end{align*} 
where $\operatorname{logits-initialization}(\cdot)$ maps each discrete token to a base, almost-one-hot logits representation in the model's vocabulary space $V$, $\{ -K, +K \}^{|V|}$. 
A 
noise schedule $\Bar{\alpha}_t$ controls the level of noise added to the base representation, where timestep ${t \sim \mathcal{U}(1, T)}$ and larger $t$ means a noisier representation. 

\methodname{}'s training loss on $\boldsymbol{w}^{c:c+B}$ is conditioned both on the noisy representation $\Tilde{\boldsymbol{w}}_t^{c:c+B}$ and the prior context $\boldsymbol{w}^{<c}$ to the block.  
\begin{align*}
    & \mathbb{E}_t[
    -\log p_\theta(\boldsymbol{w}^{c:c+B} \mid \Tilde{\boldsymbol{w}}_t^{c:c+B}, \boldsymbol{w}^{<c})]  \\
    &= \mathbb{E}_t\left[
    \sum_{j=c}^{j<c+B}-\log p_\theta(w^{j} \mid \Tilde{\boldsymbol{w}}_t^{c:c+B}, \boldsymbol{w}^{<c})\right] 
\end{align*} 
The model has access to a locally bi-directional context through the noisy representation. 
In contrast, the canonical autoregressive training loss for $\boldsymbol{w}^{c:c+B}$ would be $\sum_{j=c}^{j<c+B}-\log p_\theta(w^{j} \mid \boldsymbol{w}^{<j})$, conditioned on the uni-directional left context only.

\paragraph{Decoding} 
At inference time, given a context $\boldsymbol{w}^{<c}$, \methodname{} generates a block $\boldsymbol{w}^{c:c+B}$ through an iterative denoising algorithm, backtracking the noise timesteps from $t=T$ to 1. Each iteration $t$ consists of three main steps: (1) predict logits representation $\boldsymbol{w}_{\text{logits}, t}^{c:c+B}$ for the decoding text block using the learned model, (2) 
project the logits to an almost-one-hot representation $\hat{\boldsymbol{w}}^{c:c+B}_t$ in the base space $\{ -K, +K \}^{|V|}$ (with optional modifications), (3) add a Gaussian noise corresponding to timestep $t-1$ to the projected representation and proceed to the next iteration.
\begin{align*}
    &\boldsymbol{w}_{\text{logits}, t}^{c:c+B} = \operatorname{logits}_\theta(\boldsymbol{w}^{c:c+B} \mid \Tilde{\boldsymbol{w}}_t^{c:c+B}, \boldsymbol{w}^{<c}) \\
    &\hat{\boldsymbol{w}}^{c:c+B}_t = \operatorname{logits-projection}(\boldsymbol{w}_{\text{logits}, t}^{c:c+B}) \\
    &\Tilde{\boldsymbol{w}}_{t-1}^{c:c+B} = \sqrt{\Bar{\alpha}_{t-1}} \hat{\boldsymbol{w}}^{c:c+B}_t + \sqrt{1-\Bar{\alpha}_{t-1}} \boldsymbol{z}
\end{align*} 
where $\theta$ is the trained diffusion model and $\Tilde{\boldsymbol{w}}_T^{c:c+B}$ is initialized with a Gaussian noise.


\algrenewcommand\algorithmicindent{0.5em}%
\begin{figure*}[t]
\begin{minipage}[t]{0.495\textwidth}
\begin{algorithm}[H]
  \caption{Training (at a given $c$)} \label{alg:training}
  \small
  \begin{algorithmic}[1]
  \vspace{.02in}
      \State $\Tilde{\boldsymbol{w}}_0^{c:c+B} = \operatorname{logits-initialization}(\boldsymbol{w}^{c:c+B})$
      \State $t \sim \mathrm{Uniform}(\{1, \dotsc, T\})$
      \State $\boldsymbol{\epsilon} \sim \mathcal{N}(\boldsymbol{0}, K^2\mathbf{I})$
      \State $\Tilde{\boldsymbol{w}}_t^{c:c+B} = \sqrt{\Bar{\alpha}_t} \Tilde{\boldsymbol{w}}_0^{c:c+B} + \sqrt{1-\Bar{\alpha}_t} \boldsymbol{\epsilon}$
      \State $r \sim \mathrm{Bernoulli}(0.5)$
      \If{$r=0$}
      \State Take a gradient descent step on
      \Statex $\qquad \nabla_\theta [\sum_{j=c}^{j<c+B} -\log p_\theta(w^{j} \mid \Tilde{\boldsymbol{w}}_t^{c:c+B}, \boldsymbol{w}^{<c})]$
      \Else
      \State With gradient calculation disabled, calculate
      \Statex $\qquad \boldsymbol{w}_{\text{logits}, t}^{c:c+B} = \operatorname{logits}_\theta(\boldsymbol{w}^{c:c+B} \mid \Tilde{\boldsymbol{w}}_t^{c:c+B}, \boldsymbol{w}^{<c})$
      \State Take a gradient descent step on
      \Statex {\tiny $\qquad \nabla_\theta [\sum_{j=c}^{j<c+B} -\log p_\theta(w^{j} \mid \Tilde{\boldsymbol{w}}_t^{c:c+B}, \boldsymbol{w}_{\text{logits}, t}^{c:c+B}, \boldsymbol{w}^{<c})]$}
      \EndIf
    \vspace{.05in}
  \end{algorithmic}
\end{algorithm}
\end{minipage}
\hfill
\begin{minipage}[t]{0.495\textwidth}
\begin{algorithm}[H]
  \caption{Decoding (at a given $c$)} \label{alg:sampling}
  \small
  \begin{algorithmic}[1]
    \State $\Tilde{\boldsymbol{w}}_T^{c:c+B} \sim \mathcal{N}(\boldsymbol{0}, K^2\mathbf{I})$
    \For{$t=T, \dotsc, 1$}
    \If{$t=T$}
    \State $\boldsymbol{w}_{\text{logits}, t}^{c:c+B} = \operatorname{logits}_\theta(\boldsymbol{w}^{c:c+B} \mid \Tilde{\boldsymbol{w}}_t^{c:c+B}, \boldsymbol{w}^{<c})$
    \Else
    \State {\tiny $\boldsymbol{w}_{\text{logits}, t}^{c:c+B} = \operatorname{logits}_\theta(\boldsymbol{w}^{c:c+B} \mid \Tilde{\boldsymbol{w}}_t^{c:c+B}, \boldsymbol{w}_{\text{logits}, t+1}^{c:c+B}, \boldsymbol{w}^{<c})$}
    \EndIf
    \State Ensemble with other models if applicable, 
    \Statex $\qquad \operatorname{all-reduce}_{\Theta, \lambda}(\boldsymbol{w}_{\text{logits}, t}^{c:c+B})$
    \State $\hat{\boldsymbol{w}}^{c:c+B} = \operatorname{logits-projection}(\boldsymbol{w}_{\text{logits}, t}^{c:c+B})$ 
    \State $\boldsymbol{z} \sim \mathcal{N}(\boldsymbol{0}, K^2\mathbf{I})$
    \State $\Tilde{\boldsymbol{w}}_{t-1}^{c:c+B} = \sqrt{\Bar{\alpha}_{t-1}} \hat{\boldsymbol{w}}^{c:c+B} + \sqrt{1-\Bar{\alpha}_{t-1}} \boldsymbol{z}$
    \EndFor
    \State \textbf{return} $\argmax \Tilde{\boldsymbol{w}}_{0}^{c:c+B}$
  \end{algorithmic}
\end{algorithm}
\end{minipage}
\caption{
Training and decoding algorithms for \newmethodname{}. The training algorithm describes the training objective at an arbitrary context length $c$. The decoding algorithm can be applied multiple rounds by appending the generation from one round to the context for the next. The decoding may stop after a fixed number of rounds or until a special end-of-sequence token is encountered.
}
\label{fig:improved_algo}
\vspace{-0.8em}
\end{figure*}

\section{\newmethodname{}} 
\label{sec:fusion}


In the age of LLMs, individual users of NLP models may often face a dilemma when they wish to customize the system with their own data. On one hand, it is difficult for user-owned devices to fit very large models and smaller models are usually not powerful enough. On the other hand, uploading their data to a commercial host of large models for finetuning or long in-context learning is expensive and also may not be desirable due to privacy risks. We aim to address this dilemma in this work by proposing a collaborative inference-time algorithm between two diffusion models: a large general-purpose model (such as ones only accessible through an API) and a small model which a user can customize (\Sref{sec:collaboration_algo}). 

We first present \newmethodname{} building on top of \methodname{} with several modifications to improve its training and decoding efficiency (\Sref{sec:method_subsec}). 
We train \newmethodname{} with a larger pretraining corpus and more parameters (ranging from 0.1B to 13B) than \methodname{} and fine-tune it to follow instructions (\Sref{sec:new_experimental_setup}). 
Next, we present how different versions of \newmethodname{} 
(general-purpose large models and user-enhanced small models) 
can be effectively interpolated at inference time, outperforming their autoregressive counterparts (\Sref{sec:collab_eval_results}).

\subsection{Algorithmic improvements over \textsc{Ssd-LM}}
\label{sec:method_subsec}
\autoref{fig:improved_algo} describes the training and decoding algorithms of \newmethodname{}. We highlight the changes in \newmethodname{} over \methodname{} below. 

\paragraph{Self-conditioning}
The core idea behind self-conditioning~\citep{Chen2022AnalogBG} is that at iteration $t$, the model takes as input not just the noised sample $\Tilde{\boldsymbol{w}}_t^{c:c+B}$,  
but also a clean output from the previous timestep $t+1$, 
$\boldsymbol{w}_{\text{logits}, t+1}^{c:c+B}$. 
This allows the model to reuse useful information in the previous prediction and focus on refining it in the current timestep, allowing convergence in fewer iterations. 
That is, for $T > t \geq 1$: 

{\small 
\vspace{-.14in}
\begin{align*}
    \boldsymbol{w}_{\text{logits}, t}^{c:c+B} &= \operatorname{logits}_\theta(\boldsymbol{w}^{c:c+B} \mid \Tilde{\boldsymbol{w}}_t^{c:c+B}, \boldsymbol{w}_{\text{logits}, t+1}^{c:c+B}, \boldsymbol{w}^{<c})
\end{align*} 
}%
More specifically, the noisy representation 
$\Tilde{\boldsymbol{w}}_t^{c:c+B}$ 
and the previous timestep prediction 
$\boldsymbol{w}_{\text{logits}, t+1}^{c:c+B}$ 
are combined before the transformer blocks of $\theta$, along with the positional embeddings and timestep embeddings as follows:\footnote{As a shorthand, we dropped the superscript for token positions $c$ to $c+B$, and use $\operatorname{sm}$ for $\operatorname{softmax}$, $\operatorname{Emb}$ for the embedding layer, and $W_{\text{diff}}$ and $W_{\text{pred}}$ for the embedding matrix for the noisy representation and self-conditioning prediction.} 
{\allowdisplaybreaks
\begin{align*}
    &\Tilde{\boldsymbol{h}}_{} = W_{\text{diff}} [\operatorname{sm}(\Tilde{\boldsymbol{w}}_t)] + W_{\text{pred}}[\operatorname{sm}(\boldsymbol{w}_{\text{logits}, t+1})] \\
    &\qquad + \operatorname{Emb}_{\text{pos}}(c:c+B) + \operatorname{Emb}_{\text{diff-time}}(t/T) \\
    &\boldsymbol{h}_{}^{<c} =  \operatorname{Emb}_{\text{ctx}}(\boldsymbol{w}^{<c}) 
    + \operatorname{Emb}_{\text{pos}}(<c) \\ &\qquad +\operatorname{Emb}_{\text{ctx-time}}(t/T) \\
    &\boldsymbol{w}_{\text{logits}, t}^{c:c+B} = \operatorname{Transformer}[\operatorname{concat}(\boldsymbol{h}_{}^{<c}, \Tilde{\boldsymbol{h}})]^{c:c+B}
\end{align*} 
}

To train the model to learn to reuse the predicted logits, we add an additional forward pass during the training phase, activated with a probability $p=0.5$. 
We predict $\boldsymbol{w}_{\text{logits}, t}^{c:c+B}$ disabling gradient back-propagation, and use it in the new cross entropy loss $-\log p_\theta(w^{j} \mid \Tilde{\boldsymbol{w}}_t^{c:c+B}, \boldsymbol{w}_{\text{logits}, t}^{c:c+B}, \boldsymbol{w}^{<c})$.

\paragraph{Removing context length sampling for efficiency} 
The original training algorithm of \methodname{} first samples a context length $c \sim \mathcal{U}(1, |\boldsymbol{w}|-B)$ for each example, encodes the context bi-directionally and computes the diffusion loss for a block of $B$ tokens following that context. 
The bi-directional encoding of the context $\boldsymbol{w}^{<c}$ cannot be shared across different context sizes $c$ for the same example. 
Moreover, when the sequence length $|\boldsymbol{w}|$ is large, a high variance in the sampled $c$ across devices in distributed training reduces the effective batch size, slowing down the training considerably. 
Therefore, in the pretraining and finetuning of \newmethodname{}, we eliminate sampling different $c$'s while equivalently modeling the same training loss as shown in \autoref{fig:improved_algo} for all $\frac{|\boldsymbol{w}|}{B}$ blocks in one data, by using a special attention mask. 
The transformer modules of \newmethodname{} encode the context $\boldsymbol{w}^{<c}$ uni-directionally while preserving the bi-directional attention for the diffusion generation block $\boldsymbol{w}^{c:c+B}$. This leads to a 2x speedup in our pilot pretraining. More details can be found in \Sref{sec:parallel_c}.

\paragraph{Sharded models across time-ranges and early stopping in decoding}

We observe that at test time \newmethodname{} often shows distinct behaviors at different timestep ranges. 
We empirically divide the number of iterations into five ranges of equal sizes. 
In the beginning of decoding ($t \in (0.8T, T]$), when the noise level is very high, there is no discernable pattern in which the model's intermediate predictions ($\argmax \boldsymbol{w}_{\text{logits}, t}^{c:c+B}$) in different iterations differ from each other. 
Larger changes often happen at $t \in (0.6T, 0.8T]$ after which the majority of the content is in place, and for $t \in (0.4T, 0.6T]$ only minor changes happen sparsely to make a grammatical correction or settle down on an uncertain word choice. 
Finally, for $t \in (0, 0.4T]$, the sequence does not update at all in most cases. 
We hence hypothesize that the first three timestep ranges require different capabilities from the model. 
In \newmethodname{}, we propose to optionally train three separate models $\theta_{(0.4,0.6)}$, $\theta_{(0.6,0.8)}$, and $\theta_{(0.8,1.0)}$ for the three ranges.\footnote{A similar setup has also been explored in image diffusion as expert denoisers \citep{Feng2022ERNIEViLG2I,Balaji2022eDiffITD}.} 
We still train a single model at pretraining to save resources and only perform this step during a final finetuning as described in \Sref{sec:new_experimental_setup}.\footnote{This setup could further be improved by considering models of different sizes for the three ranges where $\theta_{(0.4,0.6)}$ and $\theta_{(0.8,1.0)}$ could contain fewer parameters as they arguably perform simpler tasks to reduce the effective inference time. We leave it as future work.} 
We start the decoding at $t=T$ and stop at $t=0.4T$, saving 40\% of the inference computation.\footnote{We report a comparison between the decoding speed of \newmethodname{} and the original \methodname{} in \Sref{sec:decoding_speed}}

\subsection{Inference-time collaboration}
\label{sec:collaboration_algo}

As shown in \methodname{}~\citep{Han2022SSDLMSS} and prior work in other domains~\citep{Dhariwal2021DiffusionMB}, diffusion models are naturally suited to allow for controlling the properties of the model outputs by interpolating the model outputs with gradients from a control function such as a classifier. 
Follow-up studies have extended this idea to classifier-free guidance where diffusion models with and without controlling attributes can be interpolated contrastively using a weighted sum of their outputs~\citep{ho2021classifier}. 
We explore a new setup of the latter idea for enabling collaboration between two versions of \newmethodname{} where we interpolate the output logits of the models. Intrinsic to the diffusion paradigm, this interpolation is sequence-level and through many iterations it leverages benefits of the bi-directional context. 

\paragraph{Setup}
We first define a \emph{core} model $\theta_{\text{core}}$ which is computationally expensive to train or deploy (e.g., a large model which can only be loaded on mutiple GPUs). We assume the model is good at general-domain instruction following. We then define a \emph{user} model $\theta_{\text{user}}$ which is computationally friendly for a typical user to run on their personal device or a cloud device to their control. 
It allows incorporating data of their specific interest which they may not prefer to input to the large model. 
For both the core and user models, we also assume they do not have access to each others' model parameters. 

We also assume a prompting instruction $\boldsymbol{w}_{\text{inst}}$ which both the models have access to, and expert data $D_{\text{user}}$ that only the user model and not the core model has access to (see \autoref{fig:fusion_fig}). 
During inference, 
\begin{compactitem}
    \item $\theta_{\text{core}}$ only takes in the prompt $\boldsymbol{w}_{\text{inst}}$, $f_{\theta_{\text{core}}}(\boldsymbol{w}_{\text{inst}})$.
    \item $\theta_{\text{user}}$ can be finetuned with $D_{\text{user}}$, or use $D_{\text{user}}$ in in-context learning. In this work, we experiment with the latter setup, where the user model takes in both the user expert data and the instruction as input, $f_{\theta_{\text{user}}}(D_{\text{user}}, \boldsymbol{w}_{\text{inst}})$.
    \item Additionally, we assume the model size $|\theta_{\text{core}}| \gg |\theta_{\text{user}}|$ (the size difference is 100x in our experiments). 
\end{compactitem}
We will discuss the specific instantiation of the setup in \Sref{sec:collab_eval_results}. In the section below, we first introduce a prominent collaboration algorithm when $\theta_{\text{core}}$ and $\theta_{\text{user}}$ are autoregressive, and then propose a novel algorithm when the models are diffusion-based \newmethodname{}.

\paragraph{Method}

The collaboration between $\theta_{\text{core}}$ and $\theta_{\text{user}}$ is essentially an ensemble of the model outputs. 
One prominent way of approaching it is through a weighted average of the models' logits at inference time.\footnote{Training-time ensemble can be achieved through methods like parameter-averaging \citep{Li2022BranchTrainMergeEP}. 
However, it is not the focus of this work since our models have drastically different shapes and we do not assume the models have access to the parameters of other models.} 
For autoregressive LMs, this averaging can be performed at the token level where the logits are first combined and then transformed into probability distribution like a product-of-experts ensemble (e.g., \citet{liu2021dexperts}). 
\begin{align*}
    &w^c \sim p_{\text{collab}}(w^c \mid D_{\text{user}}, \boldsymbol{w}_{\text{inst}}, \boldsymbol{w}^{<c}) \\
    &= \operatorname{softmax}[(1-\lambda_{\text{user}}) \operatorname{logits}_{\theta_{\text{core}}}(w^c \mid \boldsymbol{w}_{\text{inst}}, \boldsymbol{w}^{<c}) \\
    &\qquad + \lambda_{\text{user}} \operatorname{logits}_{\theta_{\text{user}}}(w^c \mid D_{\text{user}}, \boldsymbol{w}_{\text{inst}}, \boldsymbol{w}^{<c})]
\end{align*}
We also consider an extension of this setup where 
we add a contrastive term to $\theta_{\text{user}}$ without the input $D_{\text{user}}$, to promote the pointwise mutual information between the expert data and the generation conditioned on the instruction \citep{Malkin2021CoherenceBW}.\footnote{We set the contrastive hyperparameter $\alpha=1.0$ throughout the evalution, though the results with $\alpha=0.0$ follow a similar trend.}  

{\footnotesize 
    \vspace{-.14in}
    \begin{align*}
        &w^c \sim  \operatorname{softmax}[(1-\lambda_{\text{user}}) \operatorname{logits}_{\theta_{\text{core}}}(w^c \mid \boldsymbol{w}_{\text{inst}}, \boldsymbol{w}^{<c}) \\
        &\qquad + \lambda_{\text{user}} (1+\alpha) \operatorname{logits}_{\theta_{\text{user}}}(w^c \mid D_{\text{user}}, \boldsymbol{w}_{\text{inst}}, \boldsymbol{w}^{<c}) \\
        &\qquad -  \lambda_{\text{user}} \alpha \operatorname{logits}_{\theta_{\text{user}}}(w^c \mid \boldsymbol{w}_{\text{inst}}, \boldsymbol{w}^{<c})]
    \end{align*}
}%

For \newmethodname{}, the process of generating tokens is intrinsically different from autoregressive models. 
However, since it preserves the notion of logits in its iterative decoding procedure ($\boldsymbol{w}_{\text{logits}, t}^{c:c+B}$), we propose a similar logits-averaging method for a diffusion $\theta_{\text{core}}$ and $\theta_{\text{user}}$, performing an ensemble for a block of tokens at each diffusion timestep. 

{\footnotesize 
\vspace{-.14in}
\begin{align*}
    &\boldsymbol{w}_{\text{core-logits}, t}^{c:c+B} = \operatorname{logits}_{\theta_{\text{core}}}(\boldsymbol{w}^{c:c+B} \mid \boldsymbol{w}_{\text{inst}}, \boldsymbol{w}^{<c}, \Tilde{\boldsymbol{w}}_t^{c:c+B}) \\
    &\boldsymbol{w}_{\text{user-logits}, t}^{c:c+B} = \operatorname{logits}_{\theta_{\text{user}}}(\boldsymbol{w}^{c:c+B} \mid D_{\text{user}}, \boldsymbol{w}_{\text{inst}}, \boldsymbol{w}^{<c}, \Tilde{\boldsymbol{w}}_t^{c:c+B}) \\
    &\boldsymbol{w}_{\neg \text{user-logits}, t}^{c:c+B} = \operatorname{logits}_{\theta_{\text{user}}}(\boldsymbol{w}^{c:c+B} \mid \boldsymbol{w}_{\text{inst}}, \boldsymbol{w}^{<c}, \Tilde{\boldsymbol{w}}_t^{c:c+B}) \\
    &\boldsymbol{w}_{\text{logits}, t}^{c:c+B} = (1-\lambda_{\text{user}}) \boldsymbol{w}_{\text{core-logits}, t}^{c:c+B} \\
    &\qquad +  \lambda_{\text{user}} (1+\alpha) \boldsymbol{w}_{\text{user-logits}, t}^{c:c+B} -  \lambda_{\text{user}} \alpha \boldsymbol{w}_{\neg \text{user-logits}, t}^{c:c+B}
\end{align*}
}%
The above procedure is instantiated through the operation $\operatorname{all-reduce}_{\Theta, \lambda}(\boldsymbol{w}_{\text{logits}, t}^{c:c+B})$ in \autoref{fig:improved_algo}. 
\autoref{fig:fusion_fig} describes both the autoregressive and diffusion collaboration in our setup illustratively. 
It is noteworthy that for diffusion models, this manner of collaboration is only straightforward in a simplex-based model such as \newmethodname{}. Diffusion variants proposed in the literature operating on token embeddings (\Sref{sec:related_work}) are not trivially suitable for it due to a mismatch in the models' embedding space.

\section{Experimental Setup}
\label{sec:new_experimental_setup}

\paragraph{Pretraining} 
Existing work on diffusion LMs is limited to modest model sizes below the order of 1B parameters \citep{Li2022DiffusionLMIC,Han2022SSDLMSS,Dieleman2022ContinuousDF}. For example, \methodname{} has the same size as RoBERTA-large \citep{Liu2019RoBERTaAR} with 0.4B parameters. 
It is unclear whether diffusions LMs have the ability to scale like autoregressive LMs.\footnote{In fact, \citet{Strudel2022SelfconditionedED} show for embedding-based diffusion models, scaling up the embedding dimensions may hurt the performance in certain cases.} 
To answer this question,
we pretrain three versions of \newmethodname{} with 0.1B, 2.7B, and 13B parameters, on a subset of a large corpus C4 \citep{Raffel2020ExploringTL}. 
Instead of pretraining from scratch, we initialize these models using publicly available OPT models~\citep{Zhang2022OPTOP}.\footnote{\citet{Han2022SSDLMSS} find initializing from pretrained non-diffusion models help the convergence of diffusion losses in \methodname{}.} 
We consider a maximum sequence length of 500 (up from 200 in \methodname{}), with a diffusion block size $B=25$. 
On the 13B \newmethodname{} model for our main evaluation, we first do 50K warmup steps without self-conditioning and then start a 100K-step pretraining with the full algorithm. It uses approximately 38B tokens from the C4 data in total. 
Other pretraining hyperparameters can be found in \Sref{sec:hyperparam_sec}. 
We show the pretraining losses of \newmethodname{} over time in \Sref{sec:pt_loss_appendix_sec}. 
Based on the trend of pretraining losses and the scale of our pretraining data compared to recent work,\footnote{For example, the LLaMA 13B model \citep{Touvron2023LLaMAOA} uses 1T tokens from multiple corpora including C4, whereas we use 38B tokens from C4 only.} we conjecture that our \newmethodname{} models are still considerably undertrained. 
Due to our computing budget, we leave to future work a potential continued pretraining over current \newmethodname{} models on larger and better curated data.


\paragraph{Instruction finetuning} 
While \citet{Han2022SSDLMSS} show the effectiveness of pretrained \methodname{} in general-domain text continuation, in this work, we primarily investigate the use of \newmethodname{} in downstream fine-tuning tasks, particularly on chat-style instruction following.\footnote{We make an additional change while finetuning \newmethodname{} to address end of sequence (EOS) issues in variable length sequences in the downstream datasets. Since a sequence could terminate in the middle of a diffusion block, while training, we pad the sequence with the EOS token to the nearest boundary of a diffusion block of size $B$. We do not mask this padding while computing the loss. We use the standard padding token after the last diffusion block boundary. 
At inference, if the generated text block $\argmax \Tilde{\boldsymbol{w}}_{0}^{c:c+B}$ in the final iteration contains an \texttt{EOS} token, we prune the trailing tokens after the first \texttt{EOS} token in the block.} 
We finetune the models with the \dolly{} dataset\footnote{\url{https://huggingface.co/datasets/databricks/databricks-dolly-15k}. We deliberately choose to finetune with \dolly{} because as opposed to other similar datasets (e.g. the ones used to train models like Alpaca~\citep{alpaca} and Vicuna~\citep{vicuna2023}), \dolly{} has an open-source license and does not depend on distillations from OpenAI models. } containing 15K human-collected instructions and responses \citep{dolly}. 
\dolly{} covers categories like open/closed-QA, brainstorming, and creative writing, though it may still be less powerful than the distillation-based data in terms of size and quality.\footnote{We did not explore other earlier instruction tuning data like \citet{Wang2022SuperNaturalInstructionsGV} and \citet{Longpre2023TheFC} since they align less with the chat scenario of our interest. Furthermore, such data can contain a considerable amount of questions that have a very short answer (e.g., multiple choice). We leave for future work to explore the applicability of diffusion on these datasets.} 
We finetune on 95\% of the \dolly{} data and use the rest for held-out evaluation. 
We finetune with a batch size of 384 and for 500 or 1000 steps for the 0.1B/2.7B/13B models. 
As a baseline, we finetune the autoregressive model OPT (0.1B/2.7B/13B) on \dolly{} with the same setup.

\begin{table*}[t]
    \centering
    \begin{tabular}{@{}lp{0.57in}p{0.57in}p{0.57in}p{0.57in}p{0.57in}p{0.57in}p{0.57in}@{}}
    \toprule
      & {\footnotesize $\lambda_{\text{user}}=0$} & {\footnotesize $\lambda_{\text{user}}=0.1$} & {\footnotesize $\lambda_{\text{user}}=0.2$} & {\footnotesize $\lambda_{\text{user}}=0.3$} & {\footnotesize $\lambda_{\text{user}}=0.4$} & {\footnotesize $\lambda_{\text{user}}=0.5$} & {\footnotesize $\lambda_{\text{user}}=1.0$} \\
      \midrule
      \underline{Relevance} \\
      OPT$_{\{\text{core},\text{user}\}}$ & 9.76  & 9.59  & 9.61 & 9.65  & 9.65  & 9.39 & 8.23  \\
      \newmethodname{}$_{\{\text{core},\text{user}\}}$ & 9.72  & 9.65  & \textbf{9.91} & 9.85  & 9.64  & 9.52 & 7.16  \\
      $\Delta_\text{collab}$ OPT &  & \textcolor{purple}{-0.17}  & \textcolor{purple}{-0.15} & \textcolor{purple}{-0.11}  & \textcolor{purple}{-0.11}  & \textcolor{purple}{-0.37} & \textcolor{purple}{-1.53}  \\
      $\Delta_\text{collab}$ \newmethodname{} &  & \textcolor{purple}{-0.07}  & \textcolor{cyan}{\textbf{+0.19}} & \textcolor{cyan}{+0.13}  & \textcolor{purple}{-0.08}  & \textcolor{purple}{-0.20} & \textcolor{purple}{-2.56}  \\
      \underline{Factuality} \\
      OPT$_{\{\text{core},\text{user}\}}$ & \textbf{9.64}  & 9.57  & 9.51 & 9.55  & 9.57  & 9.27 & 8.15  \\
      \newmethodname{}$_{\{\text{core},\text{user}\}}$ & 9.34  & 9.49  & 9.63 & \textbf{9.64}  & 9.56  & 9.48 & 7.26  \\
      $\Delta_\text{collab}$ OPT &  & \textcolor{purple}{-0.06}  & \textcolor{purple}{-0.12} & \textcolor{purple}{-0.08}  & \textcolor{purple}{-0.05}  & \textcolor{purple}{-0.29} & \textcolor{purple}{-1.44}  \\
      $\Delta_\text{collab}$ \newmethodname{} &  & \textcolor{cyan}{+0.14}  & \textcolor{cyan}{+0.30} & \textcolor{cyan}{\textbf{+0.31}}  & \textcolor{cyan}{+0.26}  & \textcolor{cyan}{+0.15} & \textcolor{purple}{-2.03}  \\
      \underline{Informativeness} \\
      OPT$_{\{\text{core},\text{user}\}}$ & 9.30  & 9.20  & 9.12 & 9.27  & 9.06  & 8.95 & 7.41  \\
      \newmethodname{}$_{\{\text{core},\text{user}\}}$ & 8.97  & 9.02  & 9.33 & \textbf{9.36}  & 9.06  & 8.97 & 6.38  \\
      $\Delta_\text{collab}$ OPT &  & \textcolor{purple}{-0.10}  & \textcolor{purple}{-0.18} & \textcolor{purple}{-0.03}  & \textcolor{purple}{-0.24}  & \textcolor{purple}{-0.35} & \textcolor{purple}{-1.89}  \\
      $\Delta_\text{collab}$ \newmethodname{} &  & \textcolor{cyan}{+0.05}  & \textcolor{cyan}{+0.36} & \textcolor{cyan}{\textbf{+0.39}}  & \textcolor{cyan}{+0.09}  & \textcolor{purple}{0.00} & \textcolor{purple}{-2.59}  \\
      \underline{Coherence} \\
      OPT$_{\{\text{core},\text{user}\}}$ & 9.61  & 9.47  & 9.37 & 9.44  & 9.41  & 9.13 & 7.70  \\
      \newmethodname{}$_{\{\text{core},\text{user}\}}$ & 9.41  & 9.35  & \textbf{9.65} & 9.59  & 9.25  & 9.17 & 5.84  \\
      $\Delta_\text{collab}$ OPT &  & \textcolor{purple}{-0.14}  & \textcolor{purple}{-0.24} & \textcolor{purple}{-0.17}  & \textcolor{purple}{-0.20}  & \textcolor{purple}{-0.48} & \textcolor{purple}{-1.91}  \\
      $\Delta_\text{collab}$ \newmethodname{} &  & \textcolor{purple}{-0.06}  & \textcolor{cyan}{\textbf{+0.24}} & \textcolor{cyan}{+0.18}  & \textcolor{purple}{-0.16}  & \textcolor{purple}{-0.24} & \textcolor{purple}{-3.57}  \\
      \underline{Understandability} \\
      OPT$_{\{\text{core},\text{user}\}}$ & 9.66  & 9.54  & 9.53 & 9.54  & 9.51 & 9.30 & 8.10  \\
      \newmethodname{}$_{\{\text{core},\text{user}\}}$ & 9.53  & 9.56  & \textbf{9.72} & 9.67  & 9.42  & 9.34 & 6.21  \\
      $\Delta_\text{collab}$ OPT &  & \textcolor{purple}{-0.12}  & \textcolor{purple}{-0.13} & \textcolor{purple}{-0.12}  & \textcolor{purple}{-0.15}  & \textcolor{purple}{-0.36} & \textcolor{purple}{-1.56}  \\
      $\Delta_\text{collab}$ \newmethodname{} &  & \textcolor{cyan}{+0.03}  & \textcolor{cyan}{\textbf{+0.19}} & \textcolor{cyan}{+0.14}  & \textcolor{purple}{-0.11}  & \textcolor{purple}{-0.19} & \textcolor{purple}{-3.32}  \\
      \bottomrule
    \end{tabular}
    \caption{
    Evaluation of the inference-time collaboration between the large core model $\theta_\text{core}$ and the small user model $\theta_\text{user}$. A negative impact led by $\theta_\text{user}$ to $\theta_\text{core}$ is marked in red, and a positive impact in blue. \newmethodname{} is substantially more collaborative than the autoregressive OPT baseline. 
    }
    \label{tab:fusion_result}
    \vspace{-0.4em}
\end{table*}

\section{Experiments}
\label{sec:collab_eval_results}

\subsection{Inference-time collaboration} 
As introduced in \Sref{sec:collaboration_algo}, a main focus of this work is to explore the advantages of a diffusion-based LM \newmethodname{} in a collaboration setup: interpolating the outputs of a large, general model $\theta_{\text{core}}$ and a small model $\theta_{\text{user}}$ enhanced by user expert data $D_\text{user}$.

We use the 13B-parameter \newmethodname{} finetuned with \dolly{} as $\theta_{\text{core}}$ and the 0.1B finetuned \newmethodname{} as $\theta_{\text{user}}$. We use OPT 13B and 0.1B finetuned with \dolly{} under the same collaboration setup as the autoregressive baseline. 
\dolly{}'s held-out test prompts are used as $\boldsymbol{w}_{\text{inst}}$. 
A subset of \dolly{} test examples is annotated with loosely related Wikipedia passages to support the output answers;  we use these passages as a proxy for $D_\text{user}$. 
To avoid prompts with trivial answers, we additionally constrain the test instructions to those with an original annotated response of at least 50 tokens.

Inference-time collaboration is effective if the core model $\theta_{\text{core}}$ generates better responses after collaborating with the 100x smaller but user-enhanced $\theta_{\text{user}}$. 
We investigate a range of  weights  $\lambda_\text{user}$, starting from 0 where the output of the collaboration solely depends on the large $\theta_{\text{core}}$, and gradually increasing $\lambda_\text{user}$ to incorporate more $\theta_{\text{user}}$.

\paragraph{Automatic evaluation} 
We first conduct an automatic evaluation, using state-of-the-art, production-level LMs to evaluate the quality of our models' generations, which have been shown to correlate highly with human judgments and are easier to scale \citep{Liu2023GEvalNE}. 
We use GPT-3.5-turbo to rate our models' responses to the test instructions on a scale of 10, towards the aspects of relevance, factuality, informativeness, coherence, and understandability. The specific prompting template we used is detailed in \Sref{sec:eval_template}.

\autoref{tab:fusion_result} summarizes the automatic evaluation results. We observe that when $\lambda_{\text{user}}=0$ ($\theta_\text{core}$ only, no $D_{\text{user}}$ incorporated), the OPT model finetuned with \dolly{} consistently outperforms our finetuned \newmethodname{}.\footnote{We conjecture the reason is that \newmethodname{} is undertrained as discussed in \Sref{sec:new_experimental_setup} and can have a larger domain gap w.r.t. the \dolly{} data. In \Sref{sec:no_collab_eval}, we compare the intrinsic instruction following ability of \newmethodname{} and OPT using one-shot in-context learning instead of finetuning.} 
However, for OPT, collaborating with the small user model does not improve the core model's performance any further across all considered weights. Within the experimented weighting factors, $\lambda_\text{user}$ of 0.1 to 0.3 is relatively optimal, though still leading to lower scores than without collaboration. 

In contrast, the small user model $\theta_\text{user}$ improves the core model's performance in all tested attributes in \newmethodname{}. 
With appropriate weight factors ($\lambda_\text{user}=0.2, 0.3$), the collaborated \newmethodname{} system surpasses the best OPT performance in four of the five metrics and matches the fifth. 
We highlight in \autoref{tab:fusion_result} the best absolute performance and the best performance gain due to the collaboration. 
We additionally show that when $\lambda_\text{user}=1$, the small user model $\theta_\text{user}$ alone performs worse in \newmethodname{} than in OPT. This further indicates that the observed performance gain comes from an effective collaboration rather than a significantly better $\theta_\text{user}$. 

\paragraph{Human evaluation} 
To corroborate our findings, we further perform a human evaluation comparing the outputs from \newmethodname{} and OPT under a collaborative setup. For each test prompt, we show \newmethodname{} and OPT responses with $\lambda_\text{user}$ of 0.2 to the human annotators as a randomized pair. We asked the annotators to choose the preferred response while allowing for annotating equally good responses or equally bad responses. A total of 9 annotators (graduate and undergraduate researchers in NLP, not authoring this work) made 259 human preference annotations over 94 test prompts, with each response pair receiving 1-4 annotations. 
We show in \autoref{tab:human_eval_collab} that the collaboration between \newmethodname{} $\theta_\text{core}$ and $\theta_\text{user}$ is overall more preferred by humans to the OPT models under the same setup. \newmethodname{} wins in 43 cases (45.7\%) while loses only in 25 cases (26.6\%). 
We additionally measure an average Cohen's kappa coefficient between all pairs of annotators who annotated the same subset of instances. 
We observe $\kappa$=0.31 indicating a fair agreement, especially that the task is highly subjective by nature. 

Overall, through automatic and human evaluations, we show that \newmethodname{} offers unique benefits in an interesting case of inference-time collaboration, effectively fusing a general-purpose large model and a small model enhanced by some expert data.

\subsection{Ablation study: \newmethodname{} as a standalone diffusion chat model} 
\label{sec:no_collab_eval}

In this section, we divert from our main inference-time collaboration setup and investigate the capabilities of \newmethodname{} as a standalone language model. 
We are interested in the instruction following ability intrinsic to the vanilla \newmethodname{} \emph{without} inference-time collaboration or any finetuning (like with \dolly{}). 
We compare original \newmethodname{} and OPT 13B in responding to the prompts from the Vicuna test set \citep{vicuna2023}, which include problems of open-ended question answering, creative writing, etc.\footnote{Out of the 80 Vicuna test prompts, we empirically find both models constantly fail on prompts from the math and coding categories. We therefore filter them out and keep the rest 70 test cases for our experiments.} 
We formulate the setup as a one-shot in-context learning problem. Before each Vicuna test prompt, we add one fixed, handcrafted in-context example from \citet{Zhou2023LIMALI} to help the models capture the format of the answers without changing the model parameters.

The main metric we report is the win rate from an automatic evaluation based on GPT-4 \citep{OpenAI2023GPT4TR}. 
We follow the original evaluation template as introduced in \citet{vicuna2023}, prompting GPT-4 to rate \newmethodname{} and OPT responses along with explanations. 
As additional metrics, we also compute the conditional perplexity of the responses using external language models GPT-Neo-1.3B \citep{gpt-neo} and GPT-2-large \citep{Radford2019LanguageMA}. 
While there are no gold answers to the Vicuna test prompts, we use GPT-3.5's responses as reference answers and subsequently compute a BERTScore w.r.t. them for the responses from \newmethodname{} and OPT. 
As shown in \autoref{tab:icl_chat}, we overall observe a higher win rate, lower perplexity, and higher BERTScore for our diffusion language model \newmethodname{} compared to the autoregressive OPT. 
We additionally evaluate \newmethodname{} finetuned with \dolly{} and report results in \Sref{sec:standalone_ssd2_dolly_eval}. 
We show some qualitative examples of \newmethodname{}'s generations in \Sref{sec:ssd_dolly_qual_ex}.

\begin{table}[t]
    \centering
    \begin{tabular}{@{}p{0.75in}p{0.65in}p{0.75in}@{}}
    \toprule
        \newmethodname{}$_\text{collab}$ win & Draw & OPT$_\text{collab}$ win \\
        \midrule
        43 {\footnotesize (45.7\%)} & 26 {\footnotesize (27.7\%)} & 25 {\footnotesize (26.6\%)} \\
      \bottomrule
    \end{tabular}
    \caption{
    Human preference of the outputs from the inference-time collaboration experiments, comparing the diffusion-based \newmethodname{} and the autoregressive OPT. 
    }
    \label{tab:human_eval_collab}
\end{table}

\section{Related work} 
\label{sec:related_work}

Diffusion-based language models have been receiving increasing attention as a potential alternative to autoregressive language models. We identify three main categories of diffusion language models based on how they represent discrete data like text. 
Discrete diffusion language models represent language naturally as categorical data, while the diffusion or noising steps are often formulated as transition matrices \citep{hoogeboom2021argmax,Austin2021StructuredDD,He2022DiffusionBERTIG,Reid2022DiffusERDD,Zheng2023ARD,Zhou2023DiffusionNATSD}. 
Embedding-based diffusion language models often learn a mapping between the discrete language tokens and an embedding latent space, and the diffusion process is on the embedding space via a series of Gaussian noise \citep{Li2022DiffusionLMIC,Gong2022DiffuSeqST,Dieleman2022ContinuousDF,Gao2022DifformerED,Lovelace2022LatentDF,Yuan2022SeqDiffuSeqTD,Lin2022GENIELS,Ye2023DINOISERDC,Chen2023ACA,Tang2023CanDM,Balagansky2023DemocratizedDL}. 
In this work, we focus on simplex-based diffusion language models that project discrete tokens to a simplex space and perform the diffusion process with a simple Gaussian noise \citep{Han2022SSDLMSS,Mahabadi2023TESSTS}. 
Our proposed inference-time collaboration setup is most straightforward to apply to simplex-based diffusion language models, since models with different sizes share the same simplex (vocabulary) space. 
Embedding-based models over different latent representation spaces are not suitable for a direct representation interpolation. Furthermore, to the best of our knowledge, \newmethodname{} is the first of this line of literature to pretrain and finetune a diffusion language model as a chat model, encouraging future work to compare and improve over our work. We detail additional related work in \Sref{sec:additional_related_work}.

\begin{table}[t]
    \centering
    \begin{tabular}{@{}lp{0.50in}p{0.70in}p{0.74in}@{}}
    \toprule
       & Win rate & PPL $_\text{(GPT-Neo/GPT2)}$ & BERTScore $_\text{(Precision/F1)}$  \\
       \midrule
      \newmethodname{}$_\text{13B}$ & 52.3\% & {\footnotesize 7.58 / 9.62} & {\footnotesize 85.9 / 85.2} \\
      OPT$_\text{13B}$ & 47.7\% & {\footnotesize 8.44 / 10.08} & {\footnotesize 85.3 / 84.9} \\
      \bottomrule
    \end{tabular}
    \caption{
    Original \newmethodname{} responding to Vicuna test instructions in an one-shot in-context learning setup. The win rate is computed between \newmethodname{} and OPT models using the original GPT-4 evaluation introduced in \citet{vicuna2023}. 
    BERTScore is computed for the model responses w.r.t. the generations from GPT-3.5.  
    }
    \label{tab:icl_chat}
\end{table}

\section{Conclusion} 

We present an exploratory step towards 
pretraining a large simplex-based diffusion language model \newmethodname{} and finetuning it with an open-source chat-style instruction dataset. 
In a motivated setup where large general models and small user models are to collaborate with each other at inference time, we find \newmethodname{} substantially more collaborative than its autoregressive counterparts. 
These findings show the promise of diffusion language models as an instruction-following chat model and a worthy alternative to autoregressive language models.

\section*{Limitations}

In this work, we explore a novel setup of fusing large general diffusion language models and small customizable models enhanced with user expert data. 
One limitation of the proposed fusion algorithm is that it requires a search through a range of candidate balancing factors $\lambda_\text{user}$. Furthermore, a selected balancing factor remains the same across different diffusion timesteps, which is not necessarily optimal. Future work can explore and learn an optimal, dynamic schedule of the balancing factors. 
Another limitation of diffusion language models in general is a slow decoding speed compared to autoregressive models. 
Though our proposed \newmethodname{} model already includes improvements over the original \methodname{} leading to faster decoding speed (more details in \Sref{sec:decoding_speed}), future work may further adapt methods from image diffusion models targeting specifically for efficient decoding \citep{Song2021DenoisingDI,nichol2021improved,Rombach2022HighResolutionIS,Meng2022OnDO}.

\bibliography{my_cites}
\bibliographystyle{acl_natbib}

\appendix

\section{Additional related work}
\label{sec:additional_related_work}

With autoregressive language models, various efforts have been made towards building chat-style instruction following models based on open source language models \citep{Touvron2023LLaMAOA,biderman2023pythia} to replicate strong production-level closed source counterparts \citep{Ouyang2022TrainingLM,OpenAI2023GPT4TR}. Many of such work are concurrent to ours and collect high-quality finetuning datasets by distilling prompts and responses from OpenAI models \citep{alpaca,vicuna2023,Xu2023BaizeAO}. In this work, we deliberately seek fully open source data not depending on OpenAI models and adopt the \dolly{} data \citep{dolly}. We expect our models can be further improved with future releases of more curated chat-style instruction tuning datasets \citep{Zhou2023LIMALI}.


One novel setup we explored in this work is the inference-time collaboration between a large, general-purpose diffusion chat model and small, user-specific models. Inference-time collaboration has been generally explored in autoregressive models via ensembles of logits, either in an interpolation or contrastive manner \citep{liu2021dexperts,Malkin2021CoherenceBW,Li2022BranchTrainMergeEP,Peng2022ModelEI,Li2022ContrastiveDO}. 
In diffusion models, classifier-free guidance in image generation \citep{ho2021classifier} contrastively reconstruct representations with and without a controlling attribute using a single model, whereas our work collaboratively decode with models with different sizes and inputs. 
We show an unique advantage of simplex-based diffusion language models in such inference-time collaboration compared to autoregressive language models.

\section{Eliminating the sampling of context size $c$ in training}
\label{sec:parallel_c} 
In the original training algorithm of \methodname{} \citep{Han2022SSDLMSS}, they first sample a context length $c \sim \mathcal{U}(1, |\boldsymbol{w}|-B)$, and then compute the diffusion loss for reconstructing a block of length $B$ following that context. When the sequence length $|\boldsymbol{w}|$ is large, this can lead to a drastic variance in the values of $c$. Implemented naively in a distribution training setup, this setup wastes computations, and reduces the effective batch size considerably slowing down training.  
We eliminate the sampling of the context length $c$ in \newmethodname{} by processing multiple $c$'s in parallel. 
To facilitate this, we encode the context $\boldsymbol{w}^{<c}$ uni-directionally while preserving the locally bi-directional attention for the diffusion generation block $\boldsymbol{w}^{c:c+B}$. 

More specifically, assume we have a prompt $\boldsymbol{w}^{<c_0}$ and want to form the same training objective as in \autoref{fig:improved_algo} on all of the following $n$ text blocks $\boldsymbol{w}^{c_0:c_0+nB}$. We prepare a context sequence $\boldsymbol{w}^{<c_0+(n-1)B}$ and obtain $\boldsymbol{h}^{<c_0+(n-1)B}$ as described previously in \Sref{sec:method_subsec}. We prepare a diffusion sequence $\Tilde{\boldsymbol{w}}^{c_0:c_0+nB}$ and obtain $\Tilde{\boldsymbol{h}}^{c_0:c_0+nB}$ as described previously. 
Then a forward pass of $\theta$ works as below. 
\begin{align*}
    &\boldsymbol{o}_{}^{<c_0+(2n-1)B} = \operatorname{Transformer}[\\
    &\quad \operatorname{concat}(\boldsymbol{h}^{<c_0+(n-1)B}, \Tilde{\boldsymbol{h}}^{c_0:c_0+nB}); \boldsymbol{\delta}(c_0, n, B)] \\
    & \boldsymbol{w}_{\text{logits}, t}^{c_0:c_0+nB} = \boldsymbol{o}_{}^{c_0+(n-1)B:c_0+(2n-1)B}
\end{align*} 
where $\boldsymbol{\delta}(c_0, n, B)$ is a special attention mask for the transformer model, allowing a reuse of the encoded contexts while preserving the original training loss: 

{\footnotesize
\vspace{-.08in}
\begin{align*}
    &\delta_{i,j} = \begin{cases}
    \mathds{1}_{j \le i} \text{~~if~} i < c_0+(n-1)B \text{.}\\
    \mathds{1}_{j \le c_0+kB \text{~or~} c_0+(n-1+k)B<j<c_0+(n+k)B} \\
    \quad \text{if~} c_0+(n-1+k)B < i < c_0+(n+k)B \text{,}\\
    \quad \text{for~} 0 \le k < n \text{.}
\end{cases}
\end{align*} 
}
Row $i$ of $\boldsymbol{\delta}$ indicates the attention-accessible positions for the $i$-th input token of the transformer. For example, assume the original context is $[a]$ and the target generation is in two blocks $[b,c]$ and $[d,e]$. The input sequence to the \textsc{Ssd-2} transformer model is $[a,b,c,\Tilde{b},\Tilde{c},\Tilde{d},\Tilde{e}]$, and the attention mask is: 
\begin{align*}
\boldsymbol{\delta}(1, 2, 2) = 
\begin{bmatrix}
1 & 0 & 0 & 0 & 0 & 0 & 0 \\
1 & 1 & 0 & 0 & 0 & 0 & 0 \\
1 & 1 & 1 & 0 & 0 & 0 & 0 \\
1 & 0 & 0 & 1 & 1 & 0 & 0 \\
1 & 0 & 0 & 1 & 1 & 0 & 0 \\
1 & 1 & 1 & 0 & 0 & 1 & 1 \\
1 & 1 & 1 & 0 & 0 & 1 & 1 \\
\end{bmatrix}
\end{align*} 
Comparing to regular language models, \newmethodname{} has a uni-directional encoder and locally bi-directional decoder. 
In a pilot pretraining session, we observe this change leads to a twice as fast training speed compared to the original \methodname{} on a same amount of training tokens.

\section{Pretraining hyperparameters}
\label{sec:hyperparam_sec} 

For the \newmethodname{} model of each size (13B/2.7B/0.1B), we conduct two phases of training, a warmup phase without self-conditioning and a formal phase with self-conditioning using the complete algorithm shown in \autoref{fig:improved_algo}. 
Throughout all pretraining setups, we use a max sequence length of 500, a learning rate of 1e-4, and a weight decay of 0.01. 
For the 13B \newmethodname{}, we train with a warmup batch size of 768 for 50,000 steps (19B tokens) and a formal batch size of 384 for 100,000 steps (19B tokens). For the 2.7B \newmethodname{}, we train with a warmup batch size of 256 for 100,000 steps (13B tokens) and a formal batch size of 1024 for 100,000 steps (51B tokens). For the 0.1B \newmethodname{}, we train with a warmup batch size of 2,048 for 200,000 steps (205B tokens) and a formal batch size of 2,048 for 100,000 steps (102B tokens). 
We use Nvidia V100 GPUs in distributed training, and the different batch size and number of warmup steps across different models are due to the models' memory footprint and the relative cluster traffic during our pilot pretraining. 
Future work with a dedicated group of computing resources can explore pretraining for longer to mitigate the undertraining issue mentioned in \Sref{sec:new_experimental_setup}.

\section{Pretraining losses}
\label{sec:pt_loss_appendix_sec}

\begin{figure}[h]
    \centering
    \includegraphics[width=0.45\textwidth]{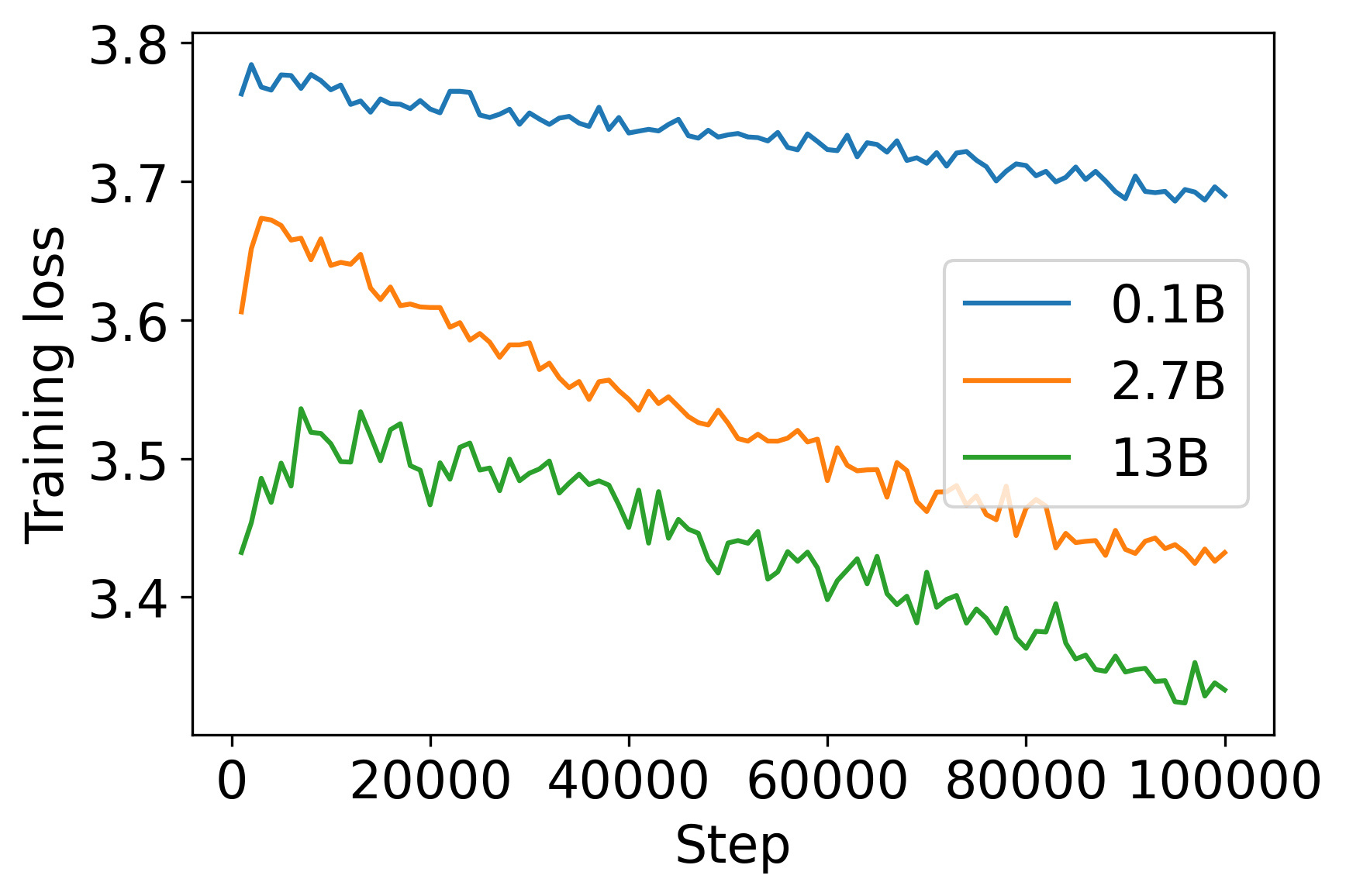}
    \caption{
    Pretraining losses across training steps (with self-conditioning, after the warmup stage). We conjecture that the models can benefit from more training given more resources. 
    }
    \label{fig:pt_loss}
\end{figure}

\autoref{fig:pt_loss} shows the pretraining losses of \newmethodname{} over time.  
We report the losses after the warmup stage and average them across batches with a self-conditioning $p=0.5$ as described in \autoref{fig:improved_algo}. 
We see a sign of undertraining from the loss curves. 
Due to our computing budget, we leave to future work a potential continued pretraining over current \newmethodname{} models on larger and better curated data. 

\section{Decoding speed} 
\label{sec:decoding_speed}
Though the decoding of \newmethodname{} is still significantly slower than an autoregressive language model, it achieves a great speedup compared to the original \methodname{}. 
We use a same setup as the reported \methodname{} decoding in \citet{Han2022SSDLMSS}. Conditioning on 50 prompting tokens, we record the speed of generating the next 25 tokens with timestep $T=1000$ on a Nvidia V100 GPU. 

The 0.4B \methodname{} takes 25 seconds. By contrast, though our 2.7B \newmethodname{} and 13B \newmethodname{} are 7x and 33x larger than \methodname{}, they only take 22 seconds and 48 seconds respectively, indicating a significant speedup.\footnote{It is worth noting that we add a timestep embedding when encoding the context in \newmethodname{} (\Sref{sec:method_subsec}). During inference across timesteps, to reuse the encoded states of the context, we quantize the input timestep $t$ to the context timestep embedding layer, $t_\text{quantized} = t - (t \mod Q)$. We use $Q=50$ throughout the work and do not find the quantization to negatively impact the generation quality empirically.}

\section{Standalone \newmethodname{} finetuned with \dolly{}} 
\label{sec:standalone_ssd2_dolly_eval}

Following \Sref{sec:no_collab_eval}, we evaluate the outputs from the finetuned models, \newmethodname{}-\dolly{} and OPT-\dolly{}, on both \dolly{}'s held-out test set and Vicuna's test set. 
As shown in \autoref{tab:head2head}, we find that against very strong baselines pretrained on much larger datasets, our model still wins on a moderate percentage of test examples. 
Compared to LLaMA (which is trained on 1T tokens for much longer but not fine-tuned for chat), \newmethodname{} performs marginally better. It is overall mildly less preferred than the OPT-\dolly{} model on both \dolly{}'s and Vicuna's test sets, and significantly less than the Alpaca model. 
We emphasize that 
compared to OPT and LLaMA-based models, \newmethodname{} is currently pretrained with a relatively small, single-corpus dataset,\footnote{Despite that we initialize \newmethodname{} with the OPT weights at the beginning of pretraining, the diffusion objective is fundamentally different from a causal language modeling objective, and the encoded knowledge of OPT may not transfer well.} and finetuned on an open-source dataset much smaller compared to its non-open-source licensed counterparts that Alpaca relies on. We believe if trained on similar datasets, \newmethodname{} can fill the current performance gap considerably.

\begin{table}[ht]
    \centering
    \begin{tabular}{@{}lp{0.4in}p{0.4in}p{0.4in}@{}}
    \toprule
      Comparing against & Win & Draw & Lose  \\
      [3pt]
      \emph{On Dolly's test set} \\
      [1pt]
      OPT-Dolly-13B & 41.4\% & 0\% & 58.6\% \\
      [3pt]
      \emph{On Vicuna's test set} \\
      [1pt]
      LLaMa-13B & 51.4\% & 0\% & 48.6\%   \\
      [1pt]
      OPT-Dolly-13B & 37.1\% & 0\% & 62.9\%   \\
      [1pt]
      Alpaca-13B & 20.0\%  & 1.4\% & 78.6\%  \\ 
      \bottomrule
    \end{tabular}
    \caption{
     Quantitative evaluation of the finetuned \newmethodname{}'s outputs against those of autogressive baselines based on ratings of GPT-4. We report the percentage our \newmethodname{} wins / draws / loses to the baseline models. 
    }
    \label{tab:head2head}
\end{table}

\section{Qualitative examples}
\label{sec:ssd_dolly_qual_ex}

In \autoref{tab:chatssd_qual_ex}, we show some qualitative examples of our finetuned diffusion language model, \newmethodname{}-\dolly{}'s outputs. The outputs are generated by the 13B model to the queries from the Vicuna test set \citep{vicuna2023}, on open-ended question answering, creative writing, etc. 
We empirically find that the generations have an overall reasonable format and content. 

\begin{table*}[t]
\begin{center}
\renewcommand{\arraystretch}{1.0}
\begin{tabular}{p{0.96\textwidth}}
    \toprule
    Prompt: How can I improve my time management skills?\\
    {\begin{lstlisting}
Time management is in short, an important skill to develop. A person needs to learn how to manage their time more effectively. Here are a few techniques that can help you do that: 
1. Prioritizing: One of the first things that you should try in order to improve your time management skills is to identify which tasks have high priority and need to be given immediate attention, and those that have low priority but can be postponed until a later date.
2. Multi-tasking: Although it might feel productive to handle more than one task at the same time, it's actually less efficient at managing our time. It's better to focus on one thing at a time, and fully dedicate 100% of our efforts to it. 
3. Delegating: It's perfectly fine to say no to some tasks or even ask for help when needed. It's sometimes considered a sign of weakness, but delegation will allow us to free up our time so that we can work on the things that really matter.
4. Being organized: Making a list will not only help us keep track of tasks that need to get done each day, but it will also help us avoid feeling overwhelmed by all the ideas and thoughts running through our heads at any given moment.
By following these tips, you'll be able to significantly improve your time management skills!
    \end{lstlisting}}\\
    \midrule
    Prompt: Write a symphony concert review, discussing the orchestra's performance and overall audience experience.\\
    {\begin{lstlisting}
It was a magical night at Symphony Hall on Saturday, as the Boston Symphony Orchestra, under the baton of Andin Mehta, performed Tchaikovsky's Symphonies No. 5 and No. 6. Tchaikovsky has always been one of my favorite composers, and these two symphonies are among his most beautiful. They are full of sweeping melodies, lyrical string writing, and lush orchestral textures that transport you to a different world each time you listen. 
And Mehta did an incredible job in leading the BSO through these masterpieces. He brought out every nuance in the music - from the quiet tenderness of the first movement to the pulsing energy of the Allegroagio to the breathtaking crescendo of the Finale. 
The entire experience was absolutely magical: I couldn't have asked for a better way to spend a Saturday night! And I would highly recommend this performance for anyone who is a fan of Tchaikovsky or classical music in general.
    \end{lstlisting}}\\
    \bottomrule
\end{tabular} 
\end{center}
\caption{
Qualitative examples of \newmethodname{}-\dolly{}'s responses to instructions. We show the first and last prompt from the Vicuna test set. The outputs of our diffusion chat model have an overall reasonable format and content, though being inaccurate in details like the conductor's name and the tempo terminology.
}
\label{tab:chatssd_qual_ex}
\end{table*}

\section{Evaluation templates} 
\label{sec:eval_template} 
We show in \autoref{tab:eval_template} the templates we used to evaluate model responses via GPT-4 or GPT-3.5-turbo. For the inference-time collaboration results, apart from the five evaluated attributes described in \Sref{sec:collab_eval_results}, in the template we additionally queried for a rating of conciseness and human preference. However, we consider conciseness a less relevant attribute compared to the other attributes and GPT-3.5-turbo less qualified to give ratings of human preference. We filtered them out in our main results but they hold a similar trend as the results of the other five attributes. 

\begin{table}[t]
\begin{center}
\renewcommand{\arraystretch}{1.0}
\begin{tabular}{p{0.48\textwidth}}
    \toprule
    Evaluation template used to compare \newmethodname{}'s responses with baseline models' responses (\Sref{sec:no_collab_eval}), following \citet{vicuna2023}. \\
    {\begin{lstlisting}
[Question]
{test instruction}

[The Start of Assistant 1's Answer]
{baseline model's response}

[The End of Assistant 1's Answer]

[The Start of Assistant 2's Answer]
{SSD-2's response}

[The End of Assistant 2's Answer]

[System]
We would like to request your feedback on the performance of two AI assistants in response to the user question displayed above.
Please rate the helpfulness, relevance, accuracy, level of details of their responses. Each assistant receives an overall score on a scale of 1 to 10, where a higher score indicates better overall performance.
Please first output a single line containing only two values indicating the scores for Assistant 1 and 2, respectively. The two scores are separated by a space. In the subsequent line, please provide a comprehensive explanation of your evaluation, avoiding any potential bias and ensuring that the order in which the responses were presented does not affect your judgment.
    \end{lstlisting}}\\
    \midrule
    Evaluation template used to rate responses from either the baseline models or \newmethodname{} w.r.t. different attributes (\Sref{sec:collab_eval_results}). \\
    {\begin{lstlisting}
Rate the response below to an instruction, from the aspects of relevance, factuality, informativeness, conciseness, coherence, understandability, and overall human preference, each on a scale of 10 (format: x/10).

========

Instruction: {test instruction}

Response: {model response}

========

Please give the ratings now.
    \end{lstlisting}}\\
    \bottomrule
\end{tabular} 
\end{center}
\caption{
Evaluation templates used in \Sref{sec:no_collab_eval} and \Sref{sec:collab_eval_results}. 
The first template was used with GPT-4 (temperature=0.2), whereas the second was used with GPT-3.5-turbo (greedy) since we need significantly more queries across different $\lambda_\text{user}$'s. 
In the comparative evaluation using the first template, flipping the order of the baseline model's response and \newmethodname{}'s response leads to a similar result. 
}
\label{tab:eval_template}
\end{table}

\end{document}